\documentclass[letterpaper, 10 pt, conference]{template/ieeeconf}
\usepackage{subfiles}
\usepackage{graphicx}
\usepackage{amsmath}
\usepackage{amsfonts}
\usepackage{multirow}
\usepackage{subcaption}
\usepackage{float}
\usepackage{tabularx}

\IEEEoverridecommandlockouts 
\ifx\secref\undefined\newcommand{\secref}[1]{Sec. \ref{sec:#1}}\fi
\ifx\figref\undefined\newcommand{\figref}[1]{{Fig. \ref{#1}}}\fi
\ifx\eg\undefined\newcommand{\eg}{{\it e.g. }}\fi

\DeclareMathOperator*{\argmin}{arg\,min}

\title{\LARGE \bf
MOTSLAM: MOT-assisted monocular dynamic SLAM\\using single-view depth estimation
}

\author{Hanwei Zhang$^{1}$, Hideaki Uchiyama$^{2}$, Shintaro Ono$^{3,4}$ and Hiroshi Kawasaki$^{5}$
\thanks{$^{1}$Graduate School of Information Science and Electrical Engineering, Kyushu University, Fukuoka, Japan}%
\thanks{$^{2}$Graduate School of Science and Technology, Nara Institute of Science and Technology, Nara, Japan}%
\thanks{$^{3}$Faculty of Engineering, Fukuoka University, Fukuoka, Japan}%
\thanks{$^{4}$Institute of Industrial Science, The University of Tokyo, Tokyo, Japan}%
\thanks{$^{5}$Faculty of Information Science and Electrical Engineering, Kyushu University, Fukuoka, Japan}%
}

\begin{document}

\maketitle
\thispagestyle{empty}
\pagestyle{empty}

\begin{abstract}

Visual SLAM systems targeting static scenes have been developed with satisfactory accuracy and robustness.
Dynamic 3D object tracking has then become a significant capability in visual SLAM with the requirement of understanding dynamic surroundings in various scenarios including autonomous driving, augmented and virtual reality.
However, performing dynamic SLAM solely with monocular images remains a challenging problem due to the difficulty of associating dynamic features and estimating their positions.
In this paper, we present MOTSLAM, a dynamic visual SLAM system with the monocular configuration that tracks both poses and bounding boxes of dynamic objects.
MOTSLAM first performs multiple object tracking (MOT) with associated both 2D and 3D bounding box detection to create initial 3D objects.
Then, neural-network-based monocular depth estimation is applied to fetch the depth of dynamic features.
Finally, camera poses, object poses, and both static, as well as dynamic map points, are jointly optimized using a novel bundle adjustment.
Our experiments on the KITTI dataset demonstrate that our system has reached best performance on both camera ego-motion and object tracking on monocular dynamic SLAM.

\end{abstract}

\section{INTRODUCTION}

Simultaneous Localization And Mapping (SLAM) states a problem that localizes the ego-motion of an agent while simultaneously building the map of an unknown environment.
By using cameras as sensors and leveraging their visual information, the above task then becomes visual SLAM.
In the past few decades, many visual SLAM methods have achieved both high robustness and performance~\cite{orbslam, lsdslam, dso}.
Especially, monocular-camera-based techniques are widely used in many robotics systems because of their great advantages in simplicity and cost-effectiveness.
However, those frameworks usually assume that scenes are static and do not deal with dynamically-moving objects.
Therefore, they cannot recognize the existence of dynamic objects.
In addition, their ego-motion estimation fails when dynamic objects exist in the scene.
Theoretically, 3D positions of dynamic features cannot be computed with monocular-camera-based motion-stereo triangulation.
With increasing applications of SLAM in various fields, such as Augmented Reality (AR), Virtual Reality (VR), and autonomous driving, the capability of understanding surrounding dynamic objects in the scene has significantly become more essential.

\begin{figure}[t]
\centering
\includegraphics[width=\linewidth]{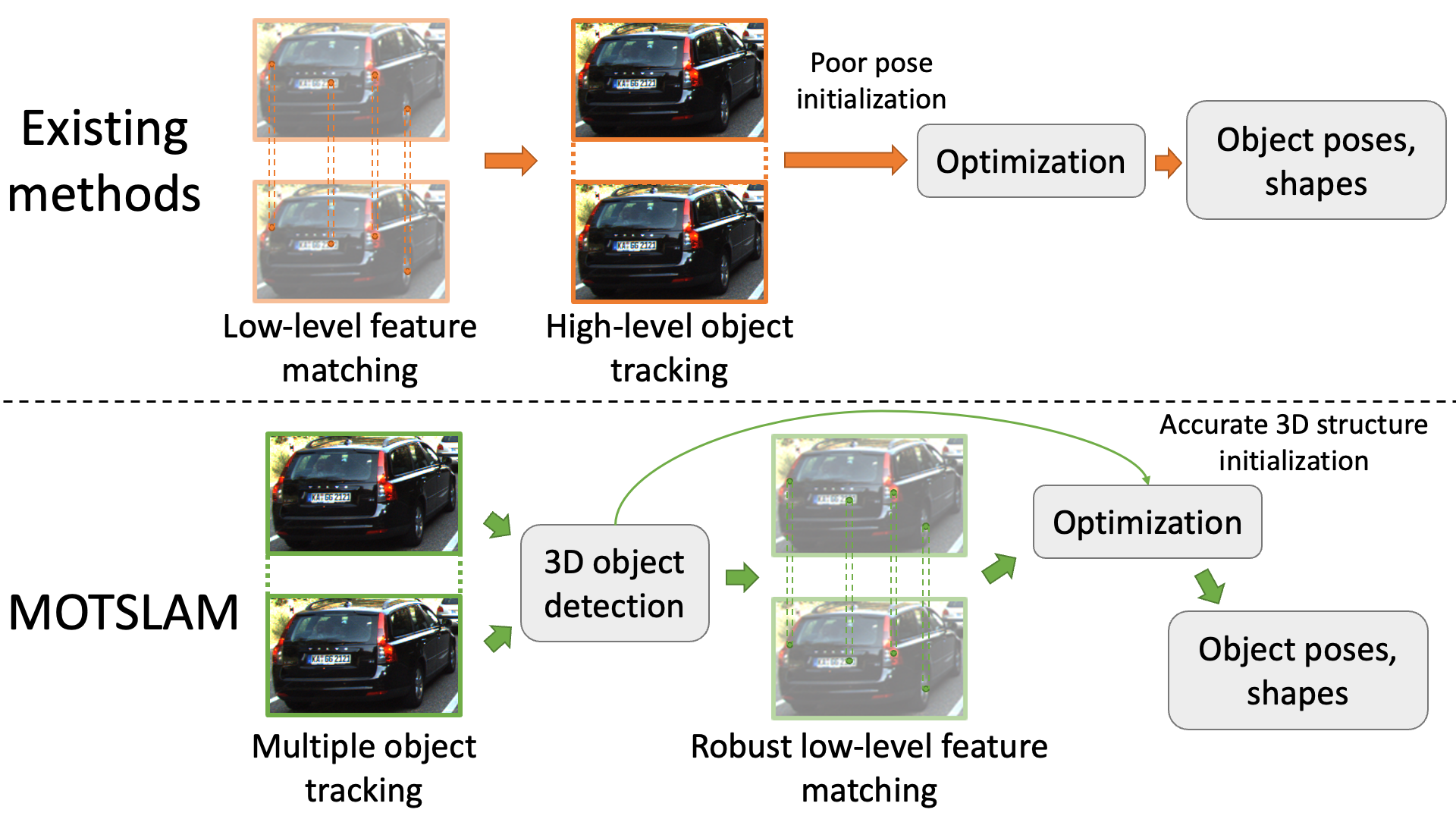}
\caption{Overview of our contributions. We propose a monocular dynamic SLAM, MOTSLAM, that performs MOT as a high-level object association to achieve robust low-level dynamic feature association. We use 3D object detection to acquire the accurate initialization of 3D object structure, which is optimized in the backend bundle adjustment.}
\label{fig:quickview}
\end{figure}

Recently, combinations of visual SLAM algorithms with deep learning techniques such as object detection and semantic segmentation have brought new possibilities to deal with dynamic objects.
In the literature, one approach is to detect possible dynamic objects and remove feature points on them as outliers during the whole procedure~\cite{dsslam,dynamicslam,dynamicdso,dynaslam}.
Another one is to simultaneously track moving objects and perform camera tracking as well as mapping~\cite{cubeslam, dynslam, lidynamic, maskfusion, dynamicfusion, gokul2020iv, clustervo}, as referred to as dynamic SLAM.
Existing methods for the latter purpose mostly adopt stereo or RGB-D configurations to directly acquire dynamic 3D structure, which cannot be achieved under the monocular setup.
Furthermore, sparse 3D features cannot provide enough clues for estimating dynamic objects.
Particularly, an insufficient number of features may degrade the accuracy of estimated poses and shapes of dynamic objects due to the difficulty of data association.

In this paper, we propose MOTSLAM, a dynamic visual SLAM with monocular frames that can track 6-DoF camera poses as well as 3D bounding boxes of surrounding objects without any additional prior for motions and objects.
To tackle the aforementioned problems, we propose to use monocular depth estimation to solve the 3D ambiguity under the monocular setup.
For the accurate and robust 3D structure of dynamic objects, we then incorporate multi object tracking~(MOT) into our framework.
Especially, the detected 2D and 3D bounding boxes are associated between frames via a 2D-based MOT technique.
In other words, our MOTSLAM efficiently combines several deep learning techniques, including 2D/3D object detection, semantic segmentation, and single-view depth estimation.
We first use deep monocular depth for possible dynamic features.
Then, we apply MOT based on 2D detection, which provides high-level association first and makes the low-level association of feature points more simple and robust even for non-consecutive frames.
This is different from the existing methods that track objects according to associated dynamic features, as illustrated in Fig.~\ref{fig:quickview}.
Furthermore, the objects tracked by MOT with associated 3D detection are initialized robustly with accurate poses and shapes.
They can be quickly optimized by using our proposed object bundle adjustment, while existing methods utilize associated 3D features to create initial states, which performance may be easily affected by outliers.
Experiments are conducted with the KITTI~\cite{kitti} dataset on both odometry and MOT to show the effectiveness of our method compared to previous methods.

In summary, our MOTSLAM solves the ambiguity of dynamic 3D structure under the monocular configuration with single-view depth estimation.
Meanwhile, MOTSLAM obtains more accurate 6-DoF poses and shapes of surrounding objects compared to existing non-monocular methods by using MOT with 3D object detection.
Our main contributions are listed as follows.
\begin{itemize}
    \item The first visual SLAM system that can simultaneously track surrounding 6-DoF dynamic objects only with monocular frames as input without any prior on motions and objects.
    \item The proposed method performs the high-level association of objects first assisted by MOT before the low-level association of features, making low-level association have better performance and robustness.
    \item Poses and shapes of objects are accurately initialized with associated 3D object detection from MOT and are refined with backend object bundle adjustment.
\end{itemize}

In the rest part of this paper, we first introduce existing research related to dynamic visual odometry / SLAM and deep techniques used in them in \secref{related-work}.
The proposed method is then explained in detail in \secref{proposed-system}.
The comprehensive experiments compared to previous methods are presented in \secref{experiments} and we conclude our work in \secref{conclusion}.

\section{RELATED WORK} \label{sec:related-work}

\subsection{Object-aware SLAM}

Recent SLAM systems have made effort to realize the existence of objects in the environment instead of sparse point clouds.
DS-SLAM~\cite{dsslam}, DynamicSLAM,~\cite{dynamicslam}, DynamicDSO~\cite{dynamicdso}, and DynaSLAM~\cite{dynaslam} treat dynamic objects as outliers and remove them by using object detection or semantic segmentation.
Existing methods that track surrounding objects simultaneously usually use stereo or RGB-D sequences as their input to acquire 3D features.
SLAM++~\cite{slam++} uses RGB-D input, detects objects with a 3D object detector, and builds an object graph refining by the pose-graph optimization, making use of the mapping results generated by SLAM.
Both Maskfusion~\cite{maskfusion} and Mid-fusion~\cite{midfusion} also acquire RGB-D inputs, and recognize and track arbitrary multiple dynamic objects with fused semantic segmentation.

DynSLAM~\cite{dynslam} is a dynamic stereo visual SLAM system that simultaneously and densely reconstructs the static background, moving objects, and the potentially moving but currently stationary objects separately.
Li {\sl et al.}'s work~\cite{lidynamic} is a stereo system specifically designed for vehicle detection by adopting a driving kinetic model and produces poses and velocities of moving cars during the visual odometry.

ClusterVO~\cite{clustervo} and DynaSLAM II~\cite{dynaslam2} are state-of-the-art stereo-camera-based sparse visual odometry systems that can track dynamic objects.
ClusterVO~\cite{clustervo} performs a clustering algorithm to aggregate 3D features as objects, while DynaSLAM II~\cite{dynaslam2} classifies features via semantic segmentation similar to ours.
Compared with other methods, they do not need a dense mapping of the stereo frame pair and do not make any assumptions of objects.
However, their 3D reconstruction of objects depends on the associated low-level 3D features heavily.
Also, they use an identity matrix with the center of mass of classified features as initial object poses, which is not always suitable in the bundle adjustment.
On the other hand, our method performs the high-level object association first using MOT with robust 3D object detection as initial states.
This loosens the heavy dependency between low-level features and object states.
Associated low-level features help optimize the state of objects, and object reconstruction still keeps its robustness if features have bad quality or low quantity.

CubeSLAM~\cite{cubeslam} is a monocular SLAM that estimates 3D boxes of objects.
It utilizes 2D object detection, semantic segmentation, and a vanish-point-based algorithm to estimate and track 3D boxes from 2D detected boxes in frames.
Also, it tracks those 3D boxes with semantic segmentation.
However, there is one constraint that a planar car motion is assumed.
Gokul {\sl et al.}~\cite{gokul2020iv} proposes a similar system to CubeSLAM, which aims at vehicle tracking and has a more precise prior shape and motion model.

While the above systems have shown significant progress of object-aware visual SLAM systems with different sensors, we notice that the existing literature under the monocular setup usually adopts strong priors on motions and objects to deal with dynamic objects properly due to the ambiguity of depth.
To solve these problems, our proposed system is capable of tracking dynamic objects robustly and accurately without any prior by maximizing the effectiveness of deep neural networks on monocular depth estimation and MOT.

\subsection{Deep learning modules in SLAM}

With the rapid development of deep learning, the efficiency cost of incorporating deep learning into visual SLAM has been significantly decreased.
Object detection, semantic segmentation, and monocular depth estimation are commonly used tools in modern dynamic SLAM systems.

To extract the high-level information of objects, object detection and semantic segmentation are widely adopted in object-aware visual SLAM systems.
Particularly, Segnet~\cite{segnet} is utilized in DS-SLAM~\cite{dsslam} for semantic segmentation, Mask r-cnn~\cite{maskrcnn} and YOLOv3~\cite{yolov3} are adopted in DynSLAM~\cite{dynslam} and ClusterVO~\cite{clustervo}, respectively.

Besides, monocular depth estimation has shown a significant step forward with deep neural networks, which can be introduced in monocular visual SLAM to acquire essential depths.
CNN-SLAM~\cite{cnnslam} fuses predicted depth maps with depth measurements of SLAM, resulting in improved quality of both estimated depth and camera ego-motion.
It adopts FCRN~\cite{fcrn} as its depth estimation method and improves the depth map by fusion as well as refinement during the SLAM procedure.
D3VO~\cite{d3vo} incorporates deep depth, pose, and uncertainty with visual odometry in a self-supervised fashion, which also produces improved results in both depth estimation and visual odometry accuracy.

\section{MOTSLAM} \label{sec:proposed-system}

\begin{figure*}[t]
\centering
\includegraphics[width=\linewidth]{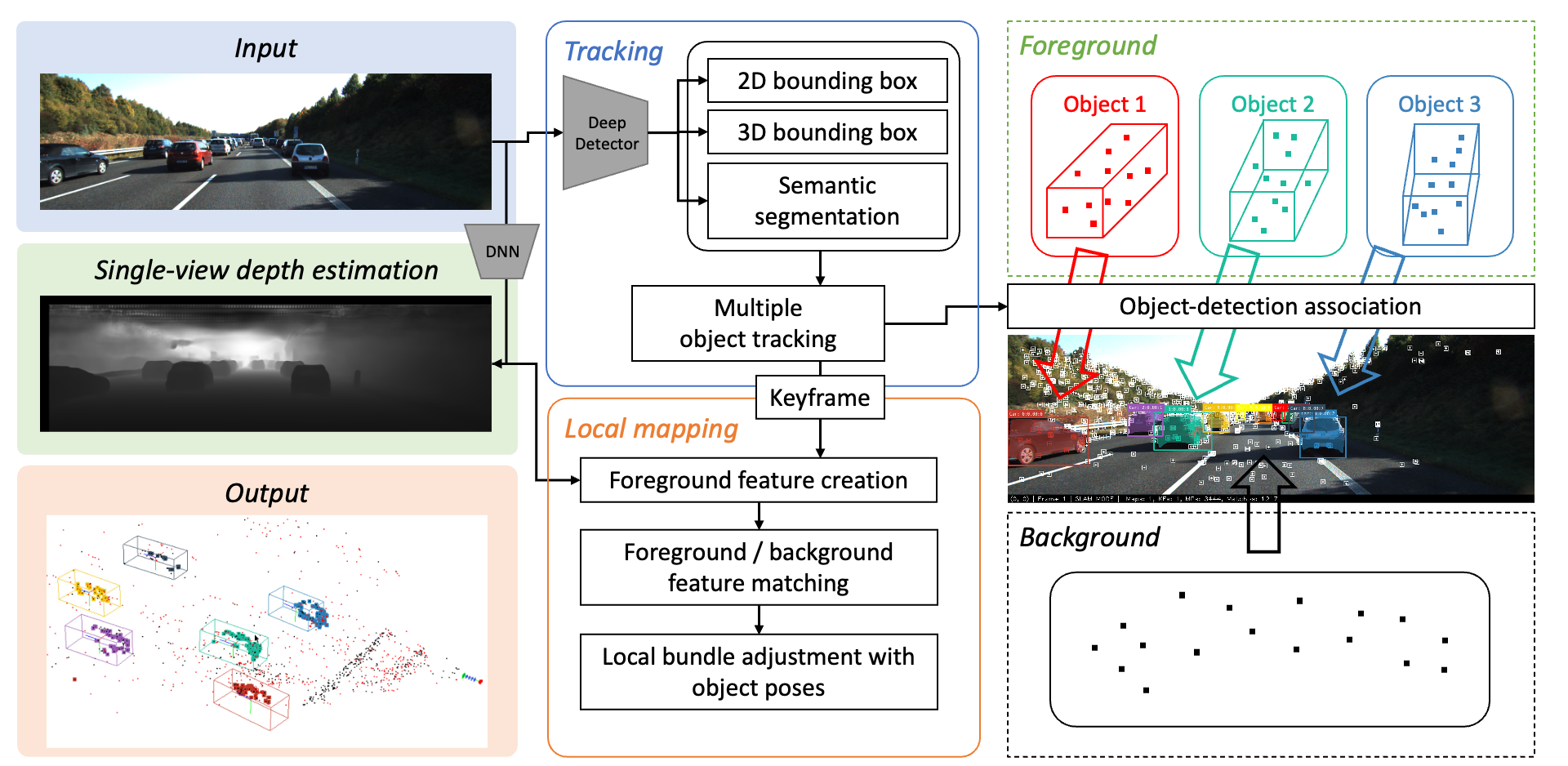}
\caption{Flow of the proposed method. Bold denotes the original pipeline of the proposed method. The system receives RGB frames solely as input. In the tracking thread, MOT is executed to track objects from the current frame. In the local mapping thread, map points in objects are created through the depth map, and are robustly associated among neighbor keyframes. Finally, the object local bundle adjustment jointly optimizes the camera pose, the current object poses, and both foreground as well as background map points.}
\label{fig:overview}
\end{figure*}

\figref{fig:overview} shows the flow of the proposed method.
MOTSLAM applies an MOT pipeline associated with object detection and semantic segmentation before local camera tracking.
Significant advantages of using MOT are that dynamic features can be simply matched by existing associated 2D object regions, and that MOT in 2D images is robust to short-term tracking lost, which means tracking can still be possibly restored when objects are missing in one or two frames.

MOTSLAM is built on top of ORB-SLAM3~\cite{orbslam3}. 
It takes monocular sequential frames as input.
For each frame, ORB~\cite{orb} extraction is executed along with 2D/3D object detection, semantic segmentation, and single-view depth estimation.
MOT is then applied to create new objects or associate existing objects to the current detection.
If a feature has its 2D observation inside an instance, it is assigned to that instance and recognized as a foreground feature with semantic segmentation.
When 2D observations do not belong to any instance, the features are then recognized as background features.
The 3D positions of foreground features are computed by using the estimated depth map and associated across neighbor keyframes using poses of objects. 
Finally, the object local bundle adjustment jointly optimizes the pose of the camera, the poses of current objects, and the associated map points.

\subsection{Notations}

At each frame $t$, the proposed method receives an RGB frame $\mathbf{I}^t$ along with an estimated depth map $\mathbf{D}^t$, and outputs: the camera pose $\mathbf{T}_{wc}^t \in SE(3)$, which transforms points from the camera frame to the world one, the state of all detected objects $\mathbf{O}^t = \{\mathbf{T}_{wi}^t, \mathbf{d}_i^t, \mathbf{c}_i^t\}_i$, where $\mathbf{T}_{wi}^t \in SE(3)$ transforms points from the object frame $i$ to the world one, $\mathbf{d}_i \in \mathbb{R}^3$ is the dimension of object $i$, $\mathbf{c}_i \in \mathbb{R}^3$ is the center of object $i$, and the state of all features $\mathbf{x}^t = \{\mathbf{z}_j^t, \mathbf{p}_j^t, o_j^t\}_j$, where $\mathbf{z}_j^t \in \mathbb{R}^2$ is the coordinate of its 2D observation, $\mathbf{p}_j^t \in \mathbb{R}^3$ is its 3D position and $o_j \in \mathbb{N}$ is the object identifier that the feature belongs to.
Specifically, when $o_j = 0$, we denote $\mathbf{x}_j$ a feature in the background.
Besides, we denote $\{\mathbf{B}_k^t\}_k$ the 2D bounding boxes detected in frame $t$, and $\{\mathbf{S}_k^t \subset \mathbf{I}^t\}_k$ the corresponding segmentation.

\subsection{Local map tracking with MOT}

After ORB features are extracted, we perform object detection and semantic segmentation through deep neural networks.
2D/3D bounding boxes and 2D masks for each bounding box are extracted and 2D bounding boxes are tracked across frames by the following MOT module.
Concretely, existing objects are associated to current 2D detection as much as possible, while the rest 2D bounding boxes are treated as new objects.

With MOT~\cite{sort}, for each existing object $\mathbf{O}_i^{t-1}$, we first predict its 2D bounding box $\hat{\mathbf{B}}_i^{t}$ at the current frame $t$ by a Kalman filter.
The state of the Kalman filter is modeled as:
$$
\mathbf{x} = [u, v, s, r, \dot{u}, \dot{v}, \dot{s}]^T
$$
where $u$ and $v$ present the coordinate of the center of the 2D bounding box, $s$ is the scale and $r$ is the aspect ratio.
The aspect ratio is assumed that it does not change over time.

With the predicted set $\{\hat{\mathbf{B}}_i^{t}\}_i$, we perform an assignment between the predicted bounding boxes and the new detected bounding boxes $\mathbf{B}_i^{t}$ by the Hungarian algorithm~\cite{sort,hungarian}.
The element of the cost matrix is defined as 1 - Intersection-over-Union (IoU) between the two bounding boxes.

Since the cost matrix is not guaranteed to be square, the number of the predicted bounding boxes $|\{\hat{\mathbf{B}}_i^{t}\}_i|$ and the number of detected bounding boxes $|\{\mathbf{B}_i^{t}\}_i|$ may differ at each frame.
When $|\{\hat{\mathbf{B}}_i^{t}\}_i| > |\{\mathbf{B}_i^{t}\}_i|$, which means some existing objects in frame $t-1$ cannot be found in $t$, we increase the age of those objects by 1.
If the age of an object is larger than a threshold $\delta$, which means it fails to be found sequentially in $\delta$ frames, this object is erased from the map.
When $|\{\hat{\mathbf{B}}_i^{t}\}_i| < |\{\mathbf{B}_i^{t}\}_i|$, some objects in frame $t$ are newly detected.
We then create new objects and assign the properties of the detection to them.

After the objects are successfully processed, we can give each 2D observation $\mathbf{z}_i^t$ a corresponding label $o_i^t \in \mathbb{N}$ according to the segmentation $\mathbf{S}_j^t$ of each object $j$.
During the local map tracking stage, we perform pose optimization with only static features and remove all features that are classified to any object.
By removing any possible dynamic feature, the accuracy of the local map tracking is expected to be improved with static features.

\subsection{Separate map point creation}

Although~\cite{orbslam3} has its strategy to detect outliers, we observed that it was likely to create ``static'' map points on that object when an object moved slowly, \eg a vehicle that moved in the same direction.
Since the triangulation does not work for dynamic features, these map points become inaccurate and will lead to errors in optimization.
To this end, we separate the map point creation process for background and foreground features instead of creating all map points with the restriction of epipolar constraints for robustness in~\cite{orbslam3}.
For foreground features, we initialize their 3D positions with the estimated depth map $\mathbf{D}^t$ instead of performing triangulation.
For background features, we keep the triangulation method.

\subsection{Foreground data association}

In the proposed method, we perform foreground data association after map point creation to perform the object bundle adjustment because we do not require foreground features in local map tracking.
However, the traditional feature matching performs re-projection from the current frame to neighbor frames or vice versa, which is solely suitable for static features.
For foreground features, we leverage this to first transform the object map point to the object frame and then perform the re-projection.
We use the pose of the corresponding detected 3D bounding box here as the transformation.
3D bounding boxes are estimated via DD3D~\cite{dd3d}, which is an end-to-end network that detects 3D bounding boxes with orientation from a single image.
Furthermore, we can ideally project the map point to the correct position owing to the fixed map points in the object frame.
More specifically, given a foreground map point $\mathbf{p}_i^m$ whose $o_i^m = j$ in frame $m$, it is re-projected into frame $n$ as follows.
$$
\hat{\mathbf{z}}_i^n = \mathbf{K}\mathbf{T}_{cw}^n\mathbf{T}_{wj}^n\mathbf{T}_{jw}^m\mathbf{p}_i^m
$$
where $\mathbf{T}_{wj}^n$ and $\mathbf{T}_{jw}^m$ indicates the transformation from object $j$ to the world in frame $n$ and the transformation from world to object $j$ in frame $m$, respectively.
Through the above process, we can associate dynamic features robustly with reliable object poses.

\subsection{Object local bundle adjustment}

The proposed method performs a novel local bundle adjustment to jointly optimize the current camera pose, the current object poses, and all inlier map points, which is called object local bundle adjustment.

For background features which $o=0$, we perform the standard bundle adjustment with re-projection error as follows~\cite{orbslam3}.
$$
\mathbf{e}_{bg}(\{\mathbf{T}^t_{cw}, \mathbf{x}^t\}_{t \in F}) = \sum_{t \in F} \rho(||\mathbf{z}_i^t - \mathbf{K}\mathbf{T}_{cw}^t\mathbf{p}_i^t||^2_\Sigma)
$$
where $F$ is the local keyframe set determined by the co-visibility graph~\cite{orbslam3}, $\mathbf{x}_F$ is the set of all background features in the keyframe set $F$, $\rho$ is the robust Huber M-estimator.
For foreground features which $o > 0$, we perform the bundle adjustment with three elements, the pose of the camera, the pose of the object, and the corresponding map points, which are described as follows.
$$
\mathbf{e}_{fg}(\{\mathbf{T}^t_{cw}, \mathbf{T}^t_{wo}, \mathbf{x}^t\}_{t \in F'}) = \sum_{t \in F'} \rho(||\mathbf{z}_i^t - \mathbf{K}\mathbf{T}_{cw}^t\mathbf{T}_{wo}^t\mathbf{p}_i^o||^2_\Sigma)
$$
The keyframe set $F'$ is not the local keyframe set determined by the original co-visibility graph but is determined with the object co-visibility graph.
The keyframe is connected in the object co-visibility graph if and only if they observe at least the same object.
Another difference is that the set of map points $\mathbf{x}_{F'}$ here contains 3D positions in object frames, which are computed and saved using the object pose when they are created.

\begin{table*}[t]
    \centering
    \begin{tabular}{c|ccc|ccc|ccc} \hline
        \multirow{2}{*}{Sequence} & \multicolumn{3}{c|}{APE} & \multicolumn{3}{c|}{R.RPE} & \multicolumn{3}{c}{T.RPE} \\
        & ORB3~\cite{orbslam3} & MonoCVO~\cite{clustervo} & Ours & ORB3~\cite{orbslam3} & MonoCVO~\cite{clustervo} & Ours & ORB3~\cite{orbslam3} & MonoCVO~\cite{clustervo} & Ours \\ \hline
        20110926-0009 & {\bf 1.8770} & 2.2812 & 1.9570 & 0.0017 & 0.0025 & {\bf 0.0016} & {\bf 0.0932} & 0.1132 & 0.1001 \\
        20110926-0013 & 1.0357 & 10.8941 & {\bf 0.4180} & 0.0022 & 0.0032 & {\bf 0.0018} & 0.1384 & 0.5314 & {\bf 0.0812} \\
        20110926-0014 & 20.2488 & 1.7096 & {\bf 1.2950} & 0.0029 & 0.0021 & {\bf 0.0020} & 0.4788 & {\bf 0.0802} & 0.1355 \\
        20110926-0051 & 5.0621 & 5.3121 & {\bf 2.0405} & 0.0017 & 0.0021 & {\bf 0.0013} & 0.2094 & 0.2569 & {\bf 0.0991} \\
        20110929-0004 & 22.1048 & 4.8259 & {\bf 2.8533} & 0.0022 & 0.0039 & {\bf 0.0012} & 0.4384 & 0.1411 & {\bf 0.1121} \\
        20111003-0047 & 156.3774 & 25.7286 & {\bf 20.8944} & 0.0092 & 0.0034 & {\bf 0.0020} & 1.1775 & 0.2112 & {\bf 0.2059} \\ \hline
        Average & 34.4509 & 8.4586 & {\bf 4.9097} & 0.0033 & 0.0029 & {\bf 0.0016} & 0.4226 & 0.2223 & {\bf 0.1223} \\ \hline
    \end{tabular}
    \caption{APE (Absolute Pose Error), R.RPE (Rotational Relative Pose Error), and T.RPE (Translation Relative Pose Error) result on KITTI raw sequences. ORB3 stands for ORB-SLAM3 in RGB-D mode using our estimated depth as input. MonoCVO stands for ClusterVO~\cite{clustervo} which uses our estimated depth instead of stereo images as input.}
    \label{tab:odometry-raw}
\end{table*}

\begin{figure*}[t]
    \centering
    
    \begin{subfigure}[b]{0.235\linewidth}
        \centering
        \includegraphics[width=\linewidth]{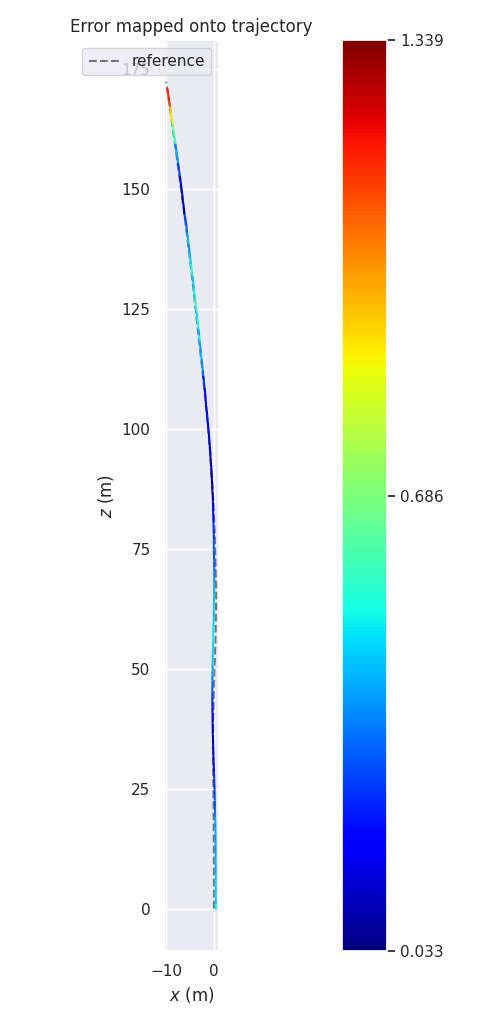}
        \caption{20110926-0013}
    \end{subfigure}
    \begin{subfigure}[b]{0.3\linewidth}
        \centering
        \includegraphics[width=\linewidth]{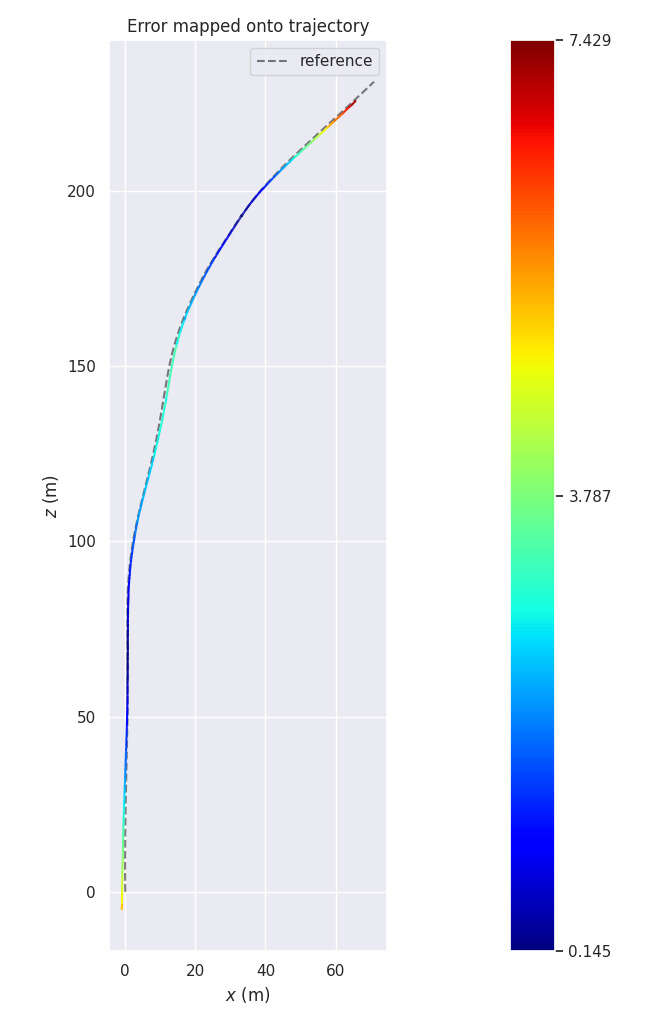}
        \caption{20110926-0051}
    \end{subfigure}
    \begin{subfigure}[b]{0.23\linewidth}
        \centering
        \includegraphics[width=\linewidth]{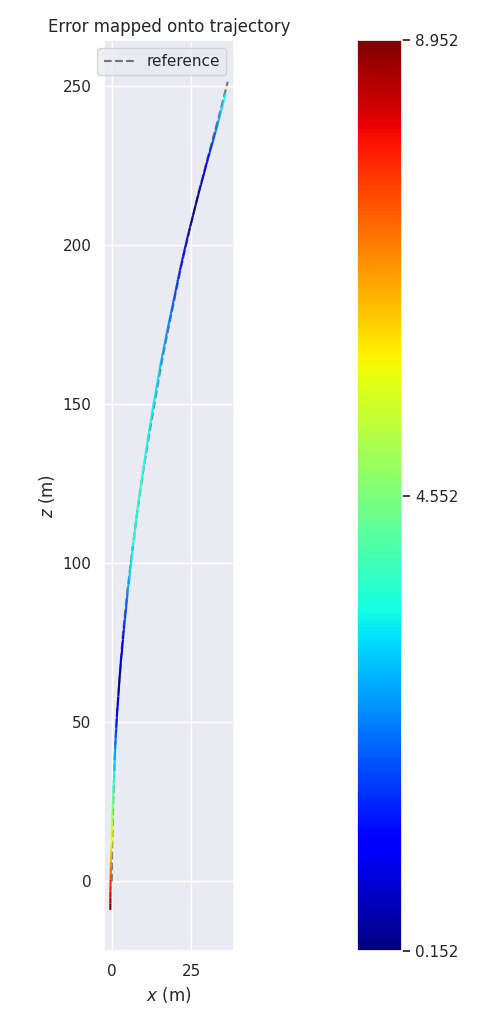}
        \caption{20110929-0004}
    \end{subfigure}
    \begin{subfigure}[b]{0.21\linewidth}
        \centering
        \includegraphics[width=\linewidth]{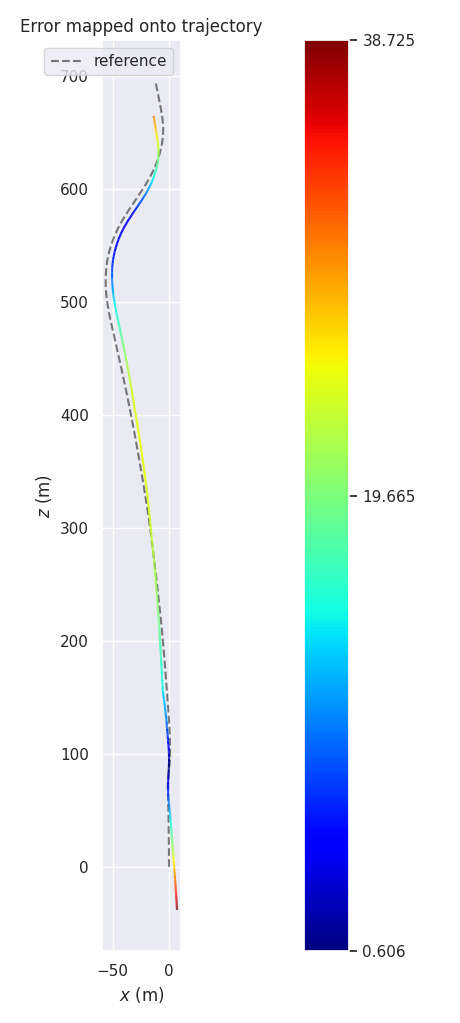}
        \caption{20111003-0047}
    \end{subfigure}
    \begin{subfigure}[b]{0.3\linewidth}
        \centering
        \includegraphics[width=\linewidth]{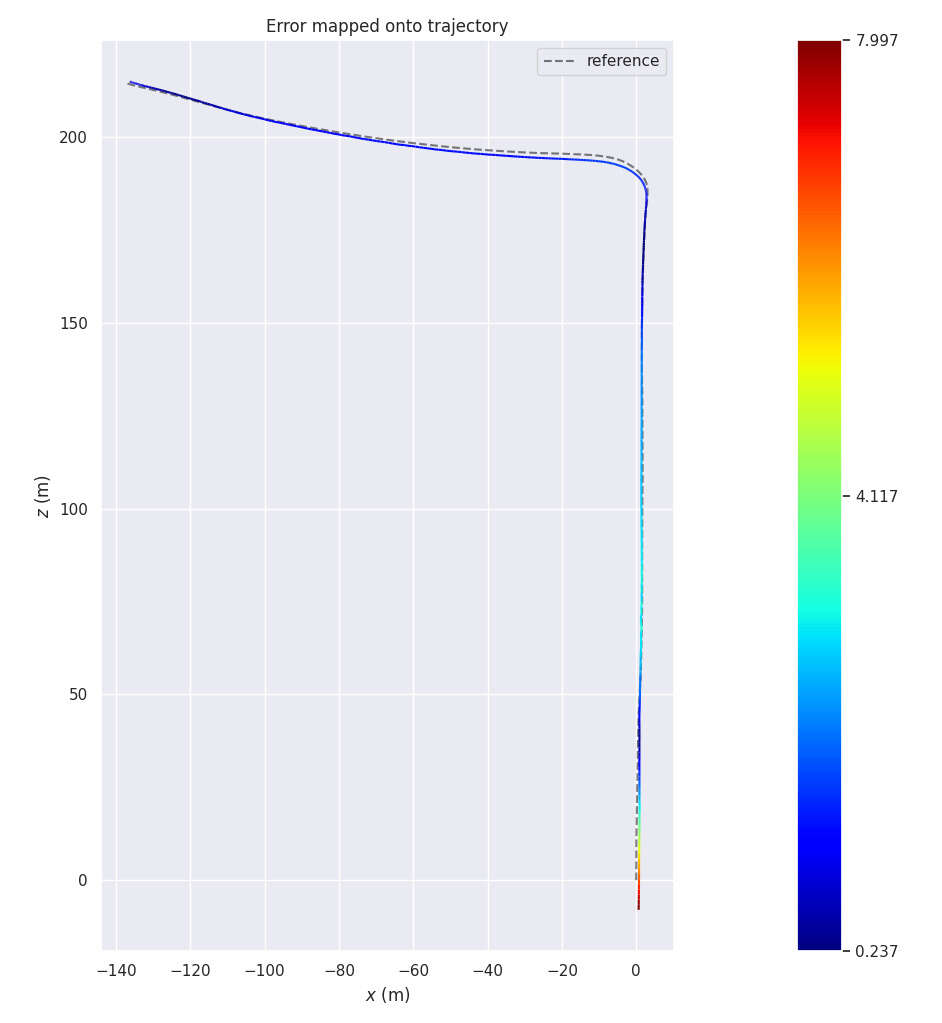}
        \caption{20110926-0009}
    \end{subfigure}
    \begin{subfigure}[b]{0.65\linewidth}
        \centering
        \includegraphics[width=\linewidth]{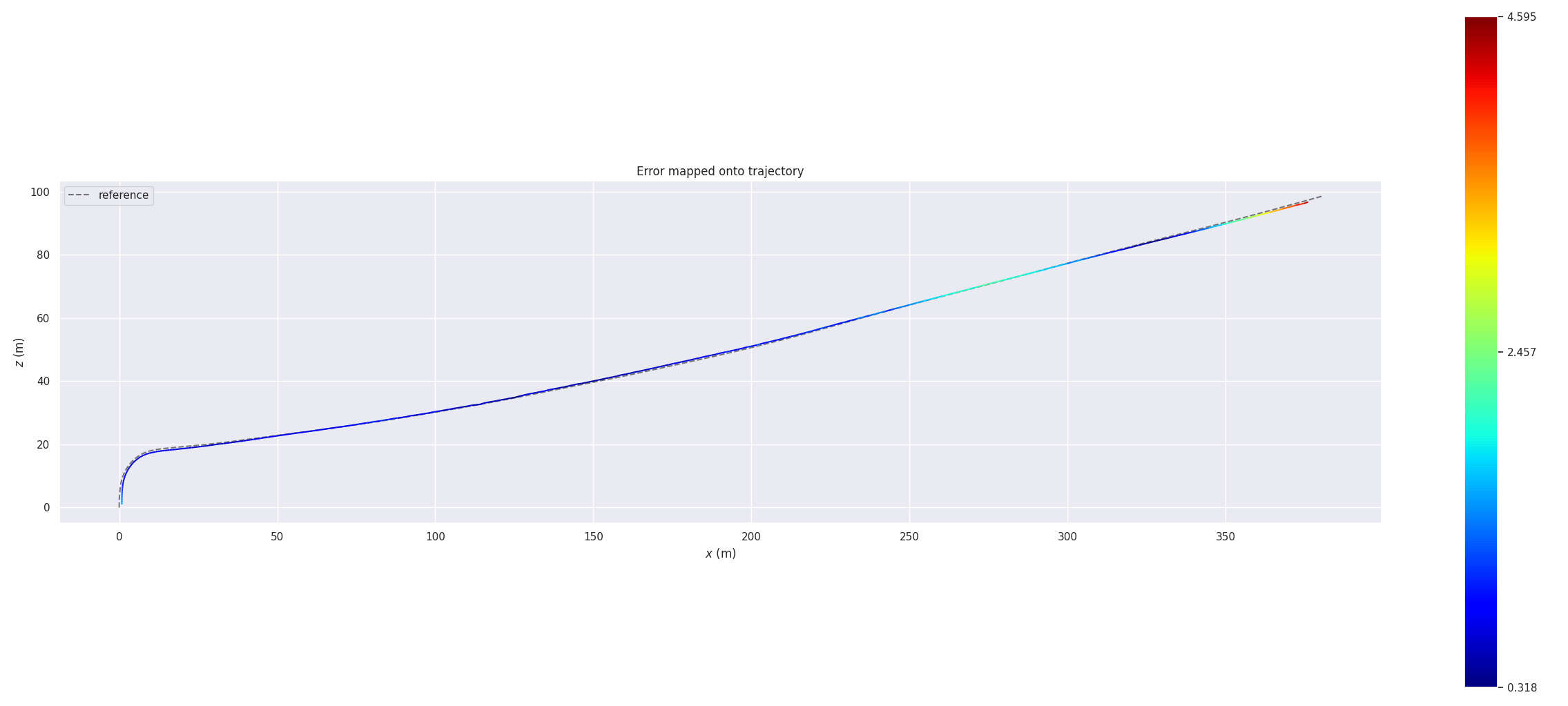}
        \caption{20110926-0014}
    \end{subfigure}
    \caption{Qualitative results of camera trajectories. The alignment algorithm in the evaluation tool~\cite{evo} is utilized to eliminate the nondeterministic scaling in the single-view depth estimation. The color represents the error comparing to the ground truth trajectory after the alignment. Since the proposed system mainly focuses on tracking dynamics, we choose sequences containing sufficient moving vehicles rather than complex long odometry tasks.}
    \label{fig:odometry-traj}
\end{figure*}

Finally, the optimization can be written as follows.
$$
\mathbf{T^*_{cw}}, \mathbf{T^*_{wo}}, \mathbf{x^*} = \argmin_{\mathbf{T_{cw}}, \mathbf{T_{wo}}, \mathbf{x}} \sum \mathbf{e}_{bg} + \mathbf{e}_{fg}
$$
where $\mathbf{T^*_{cw}}, \mathbf{T^*_{wo}}, \mathbf{x^*}$ represent the optimized camera poses, object poses and 3D feature points.
For the scenes that contain a small number of dynamic objects, the background error term takes the main effect and provides an accurate camera tracking similar to ~\cite{orbslam3}.
For the scenes that contain a large number of moving objects and few available background features, the system can still keep optimizing the camera pose with reliable object poses from 3D detection as initial values.
This guarantees not only the robustness against severe dynamic scenes but also a satisfactory accuracy of camera trajectories.


\section{EXPERIMENTS} \label{sec:experiments}

We comprehensively evaluate the proposed method performance of both camera ego-motion and object tracking to demonstrate the effectiveness of our system.
In this section, the detailed experimental results are presented.

\subsection{Datasets and setup}

We adopt the KITTI Vision Benchmark Suite~\cite{kitti} as our evaluation dataset.
We use some raw sequences that contain a significant number of objects for camera ego-motion evaluation and the MOT benchmark for object tracking evaluation.

For 2D object detection and semantic segmentation, we integrate detectron2~\cite{detectron2} into the proposed method with the Torchscript deployment.
The neural network model for 2D object detection and segmentation is Mask R-CNN~\cite{maskrcnn} with ResNet50+FPN as the backbone.
For 3D object detection, we leverage DD3D~\cite{dd3d} with its DLA-34 pretrained model.
For single-view depth estimation, we adopt BTS~\cite{bts}, which has satisfactory performance targeting the KITTI dataset.
We filter the object detection results that have confidence lower than 0.9 and merge the results of detectron2 and DD3D when two bounding boxes have a value of IoU larger than 0.8.

For evaluation tools, we use evo~\cite{evo} for camera trajectory evaluation.
This tool automatically aligns the input trajectory to the ground truth with the Umeyama algorithm~\cite{umeyama}.
For object tracking evaluation, we use AB3DMOT~\cite{AB3DMOT}, which outputs CLEAR metrics~\cite{clear} for both 2D and 3D evaluation.
We also present object detection precision using the official tool from 3D Object Detection Evaluation in KITTI~\cite{kitti}.

\subsection{Camera tracking evaluation}

We demonstrate results on raw sequences, which are chosen based on~\cite{clustervo} and contain a significant number of vehicles.
\figref{tab:odometry-raw} shows the results of absolute pose error (APE), relative pose error of rotation (R.RPE) and translation (T.RPE).
\figref{fig:odometry-traj} is the qualitative visualization of aligned trajectories by the evaluation tool~\cite{evo}.
For a fair comparison, we use the same estimated depth maps in~\cite{orbslam3} and~\cite{clustervo} as in our proposed method.
For~\cite{orbslam3}, we use the RGB-D mode with our estimated depth maps as input.
For~\cite{clustervo}, they originally use stereo images as their input.
We replace the depth inputs from stereo with our estimated depth maps.

For all 6 sequences, the proposed method reaches the best results in APE for 5 sequences.
Only for Sequence 20110926-0009, our system shows a slightly lower accuracy compared to \cite{orbslam3}.
We observe that Sequence 20110926-0009 contains mostly static vehicles.
When facing static objects, since the feature points in static objects are also removed from the camera tracking computation, our system has a fewer number of feature points to compute the camera trajectory.
Therefore, our system shows a slightly lower accuracy in this sequence.
However, for other sequences, since they all contain a significant number of moving vehicles, our system significantly outperforms other methods.

The main effectiveness of our system in camera ego-motion estimation is the removal of possible dynamic feature points.
For dynamically moving feature points, since the constraint in Perspective-n-Point (PnP) is violated, dynamic feature points introduce error to the pose optimization.
By removing those feature points from the optimization procedure, our method shows a significantly better performance especially when the scene contains a large number of moving objects such as in Sequence 20111003-0047.

\subsection{Object tracking evaluation}

\begin{table}[h]
    \centering
    \begin{tabular}{c|ccc|ccc} \hline
         & \multicolumn{3}{c|}{AP\textsubscript{bv}} & \multicolumn{3}{c}{AP\textsubscript{3D}} \\
         & Easy & Moderate & Hard & Easy & Moderate & Hard \\ \hline
         ~\cite{clustervo} & 74.65 & 49.65 & 42.65 & 55.85 & 38.93 & 33.55 \\
         ~\cite{lidynamic} & {\bf 88.07} & {\bf 77.83} & {\bf 72.73} & {\bf 86.57} & {\bf 74.13} & {\bf 68.96} \\
         ~\cite{dynslam} & 71.83 & 47.16 & 40.30 & 64.51 & 43.70 & 37.66 \\
         ~\cite{dynaslam2} & 64.69 & 58.75 & 58.36 & 53.14 & 48.66 & 48.57 \\ \hline
         Ours & \underline{80.58} & \underline{71.85} & \underline{62.90} & \underline{71.18} & \underline{63.56} & \underline{56.16} \\ \hline
    \end{tabular}
    \caption{Comparison of average precision (in \%) of object detection on KITTI tracking dataset. Bold and underline represent the best and second-best results, respectively. Besides ours, which applies a monocular configuration, other methods all apply stereo images as their input.}
    \label{tab:object-detection}
\end{table}

We also present quantitative and qualitative results of object tracking due to its the main novel capability of the proposed method.
\figref{tab:object-detection} demonstrates the precision of object detection comparing to state-of-the-art methods.
We compare our method with several state-of-the-art dynamic SLAM/VO systems.
Note that they all apply a stereo configuration as input, while our MOTSLAM uses a monocular setup with single-view depth estimation.
From the result, we can observe that our proposed monocular method outperforms ClusterVO~\cite{clustervo}, DynSLAM~\cite{dynslam}, and DynaSLAM II~\cite{dynaslam2} significantly.
We only fail to outperform~\cite{lidynamic}, which performs an alignment using extra 3D point cloud information.
We evaluate the result on the 21 sequences of the training set in the KITTI MOT dataset.
The overlapping threshold is set to 0.25, which is the same as the compared methods.

\begin{figure}[h!]
    \centering
    \begin{subfigure}[b]{\linewidth}
        \centering
        \includegraphics[width=0.45\linewidth]{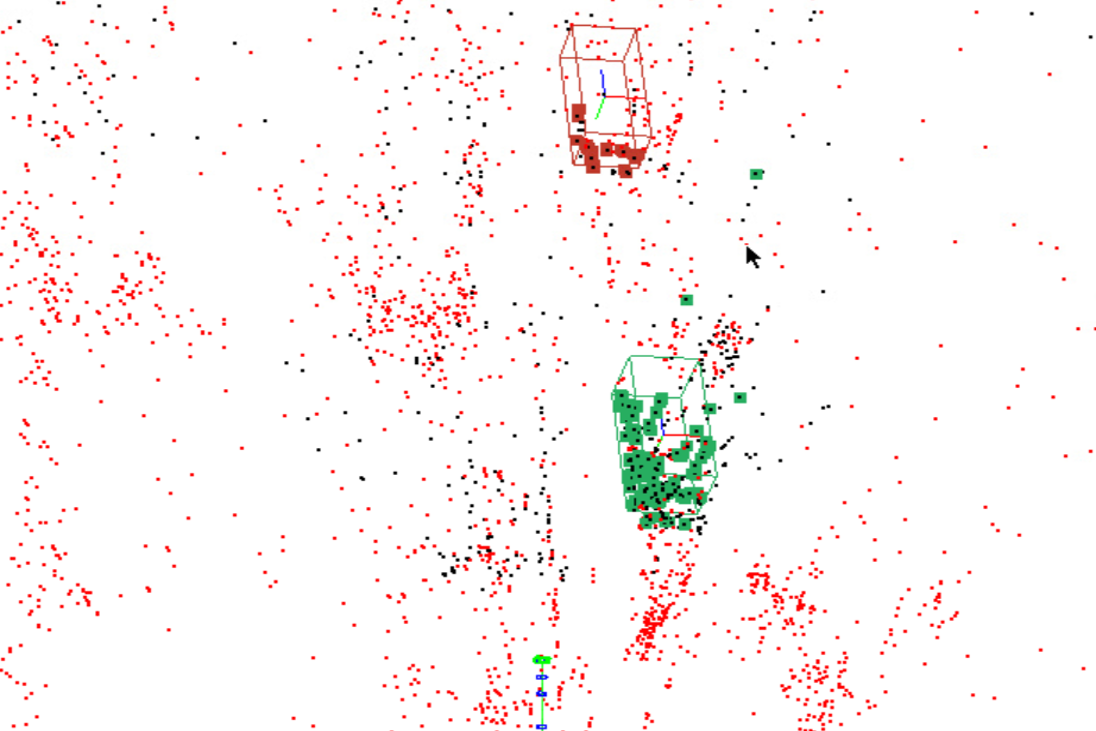}
        \includegraphics[width=0.45\linewidth]{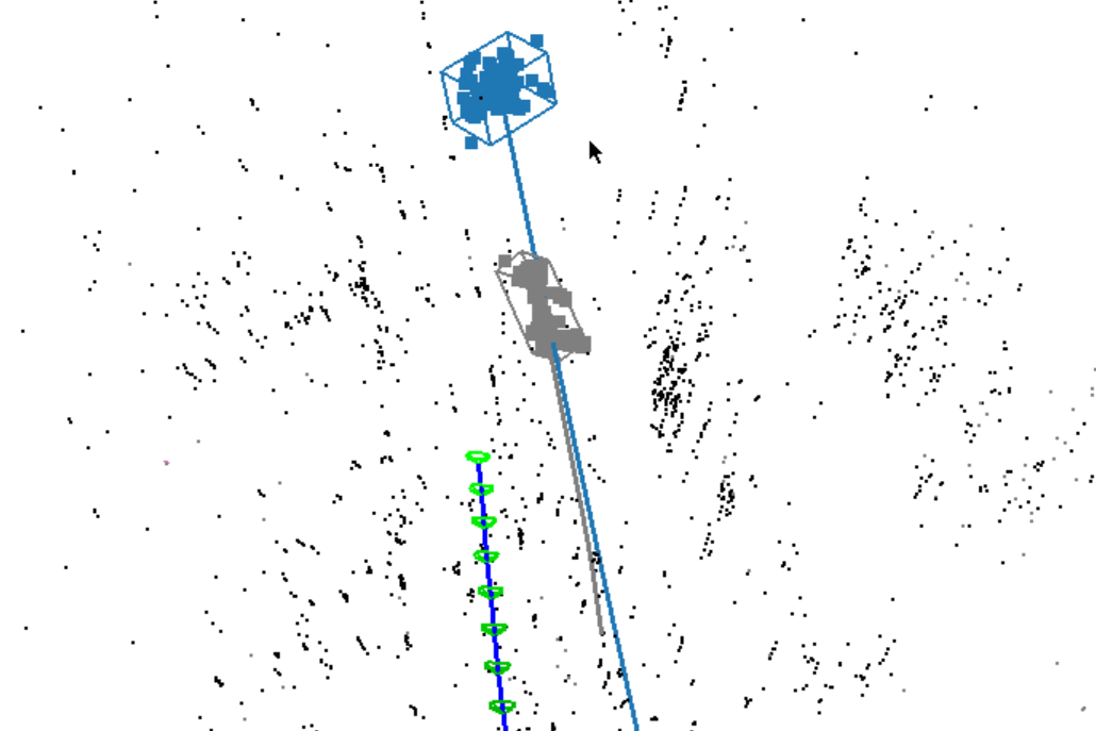}
        \includegraphics[width=0.45\linewidth]{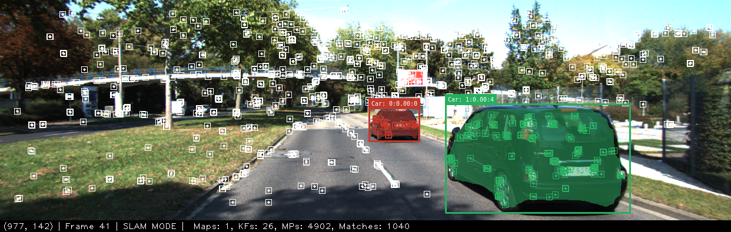}
        \includegraphics[width=0.45\linewidth]{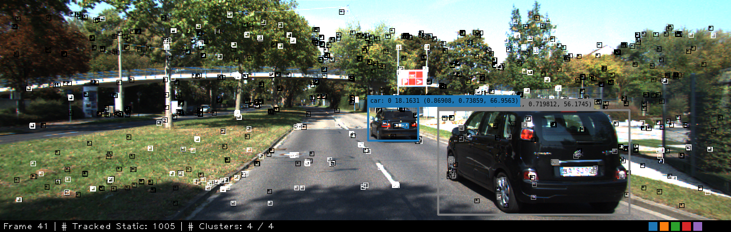}
        \caption{20110926-0013}
    \end{subfigure}
    \begin{subfigure}[b]{\linewidth}
        \centering
        \includegraphics[width=0.45\linewidth]{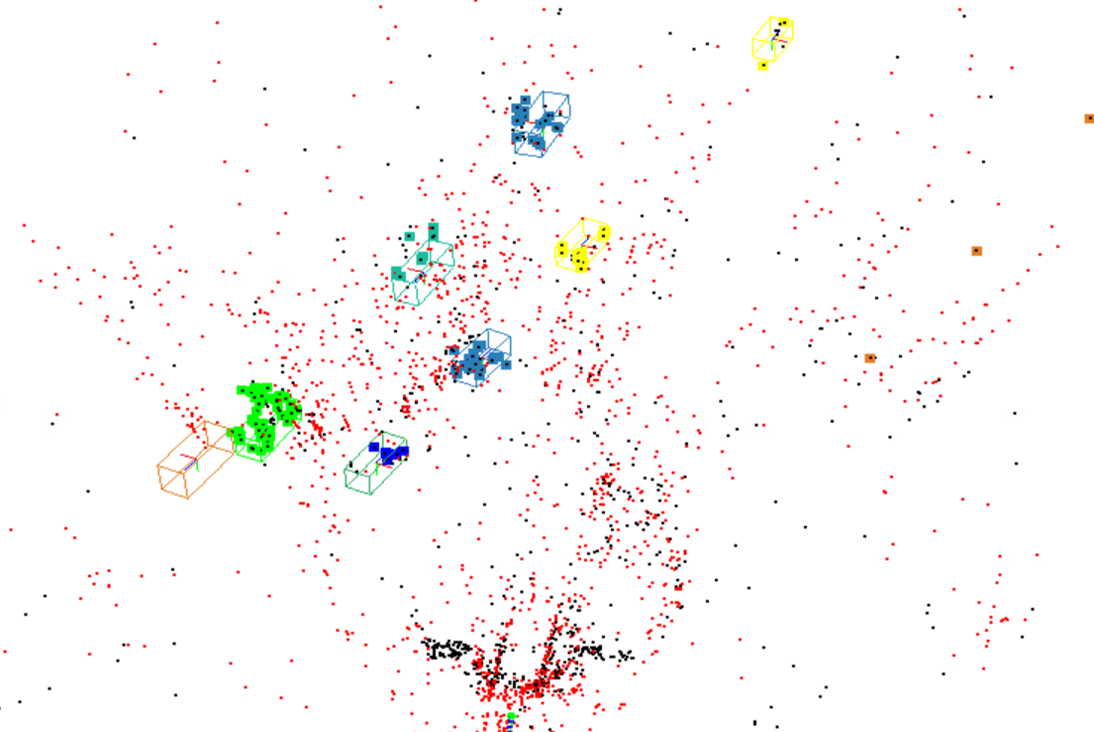}
        \includegraphics[width=0.45\linewidth]{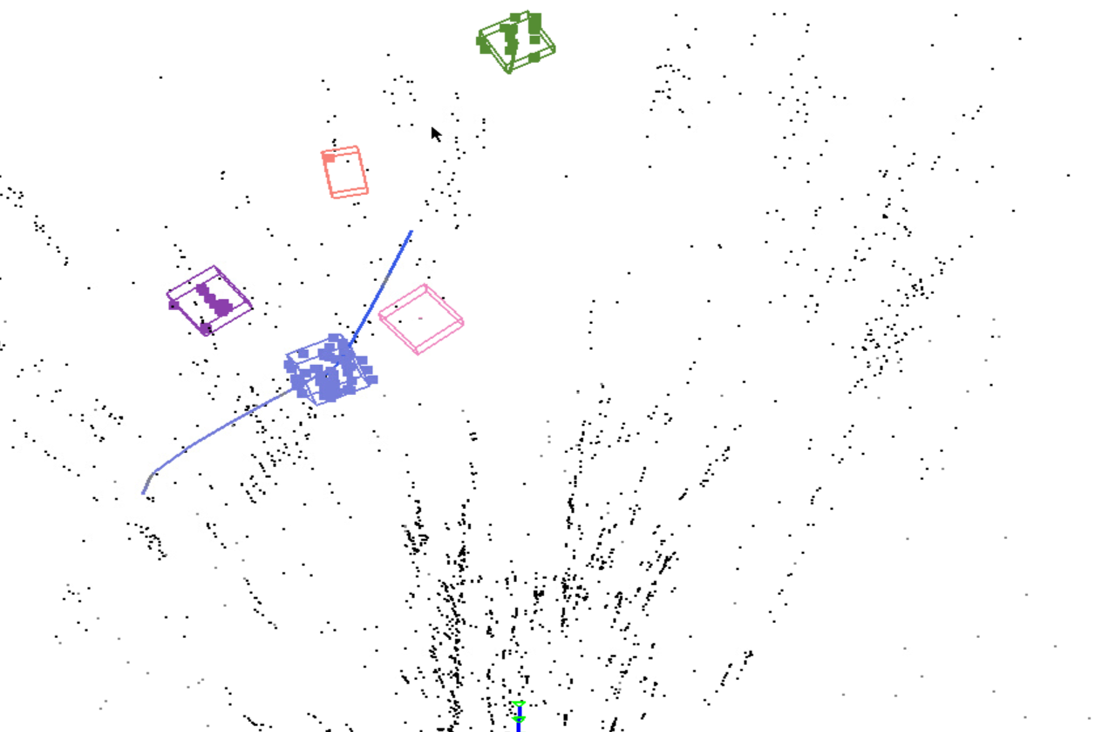}
        \includegraphics[width=0.45\linewidth]{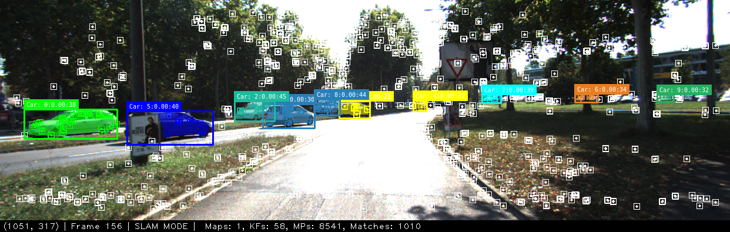}
        \includegraphics[width=0.45\linewidth]{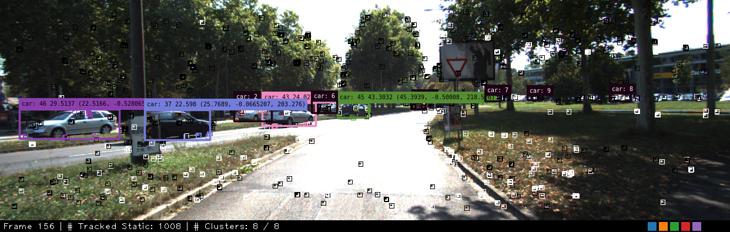}
        \caption{20110926-0051}
    \end{subfigure}
    \begin{subfigure}[b]{\linewidth}
        \centering
        \includegraphics[width=0.45\linewidth]{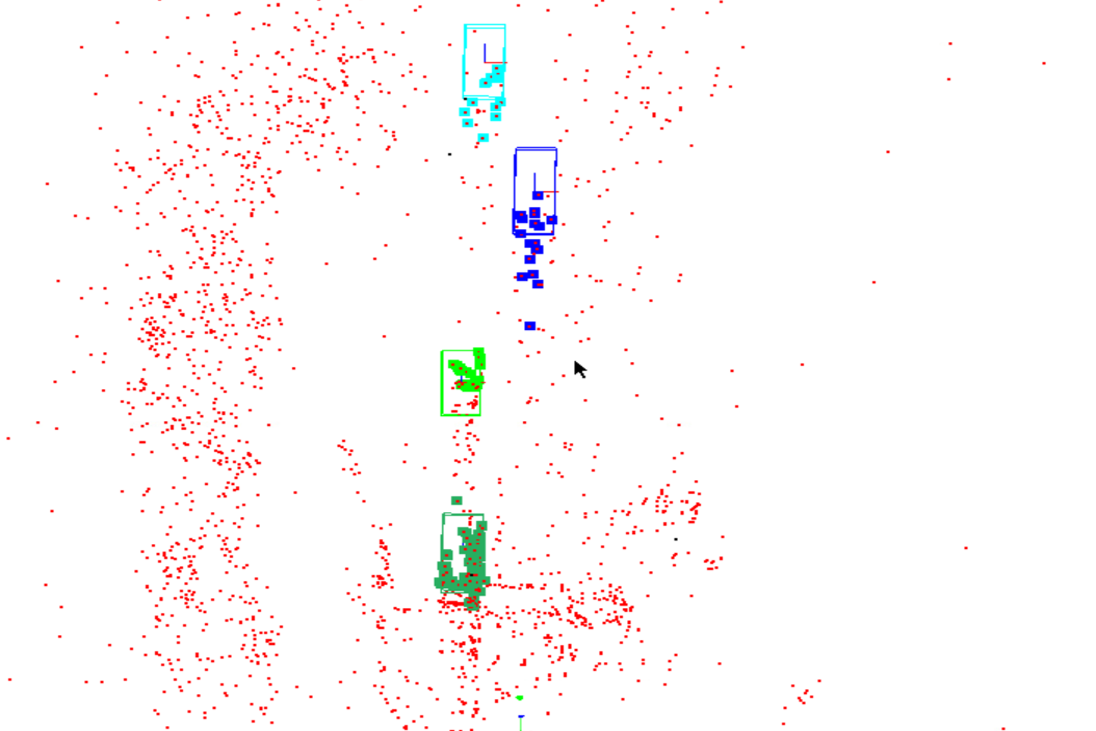}
        \includegraphics[width=0.45\linewidth]{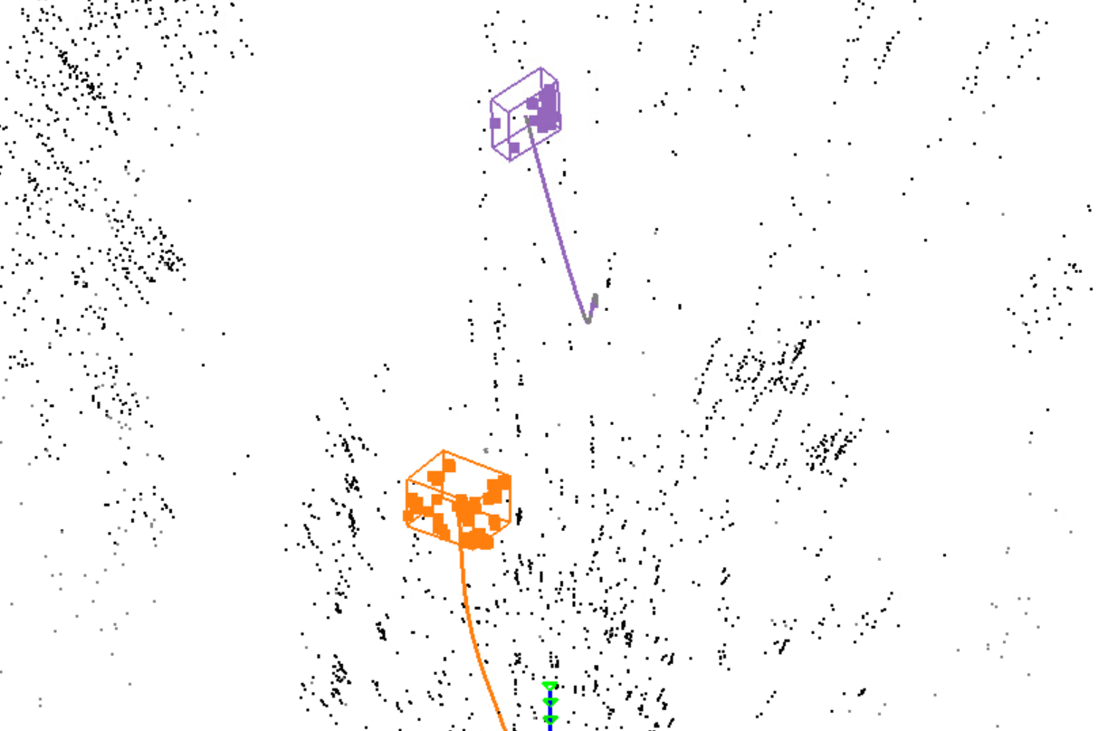}
        \includegraphics[width=0.45\linewidth]{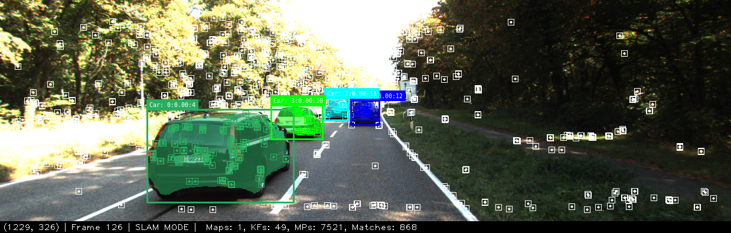}
        \includegraphics[width=0.45\linewidth]{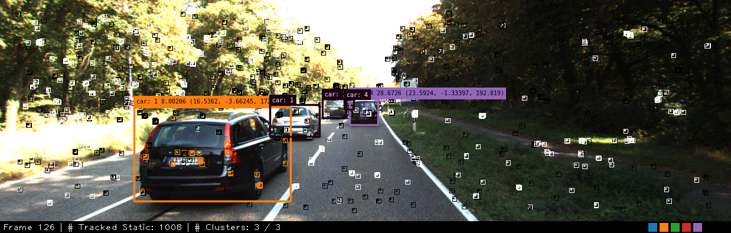}
        \caption{20110929-0004}
    \end{subfigure}
    \begin{subfigure}[b]{\linewidth}
        \centering
        \includegraphics[width=0.45\linewidth]{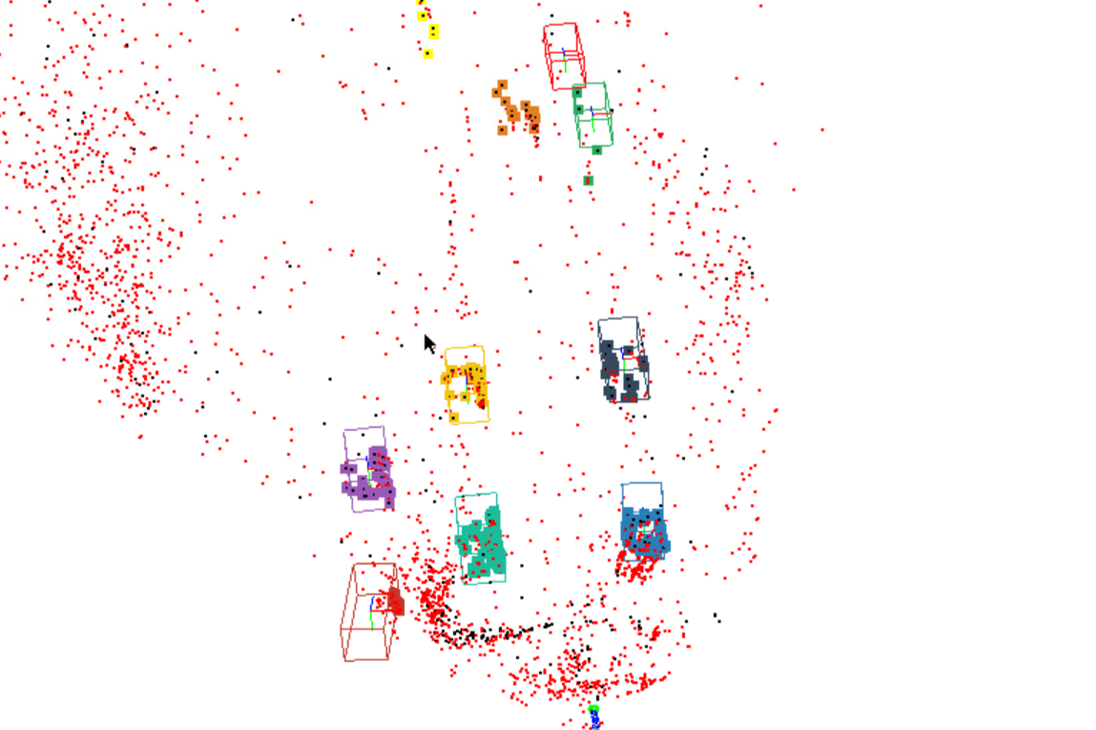}
        \includegraphics[width=0.45\linewidth]{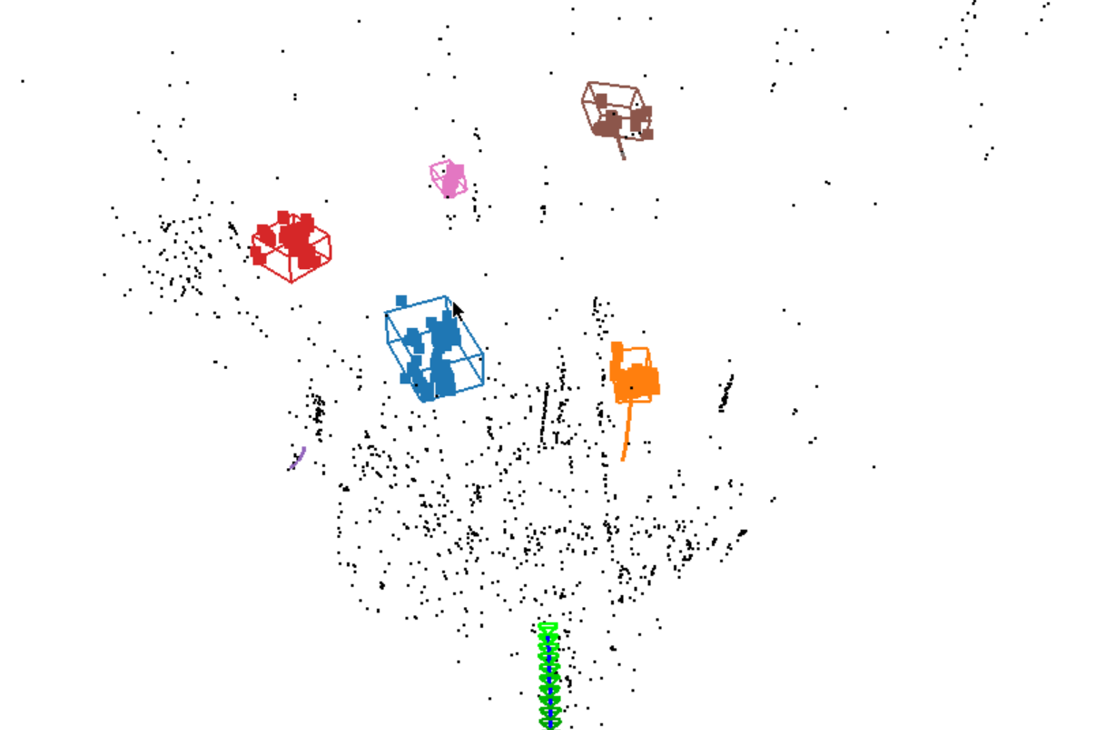}
        \includegraphics[width=0.45\linewidth]{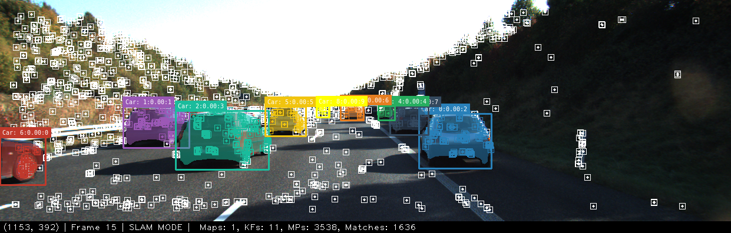}
        \includegraphics[width=0.45\linewidth]{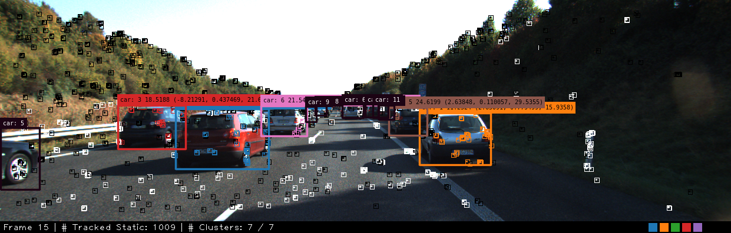}
        \caption{20111003-0047}
    \end{subfigure}
    \caption{Qualitative results of Sequence 20110926-0013, 20110926-0051, 20110929-0004, and 20111003-0047. Left is our proposed method and right is~\cite{clustervo}.}
    \label{fig:object-tracking}
\end{figure}

ClusterVO~\cite{clustervo} and DynaSLAM II~\cite{dynaslam2} estimate poses of objects via non-linear optimization using associated 3D features.
They generate 3D bounding boxes that fit corresponding features properly.
In this way, their 3D bounding boxes may fail to form a regular shape of the object.
As shown in \figref{fig:object-tracking}, MOTSLAM tends to generate more regular 3D bounding boxes of cars.
Additionally, the capability of reconstructing dynamic objects of~\cite{clustervo} and~\cite{dynaslam2} tightly depends on the quality and quantity of 3D features.
Our proposed method, however, generates initial poses and shapes of surrounding objects using 3D object detection, which loosens the dependency of 3D features and guarantees a high reconstruction performance even with insufficient features.
Although the proposed method fails to outperform~\cite{lidynamic}, it can deal with objects without specific motion models and still achieves a competitive result compared to the segmentation only method in~\cite{lidynamic}.

\figref{tab:object-tracking} shows tracking performance using the same tracking dataset.
We demonstrate the evaluated MOTA (Multiple Object Tracking Accuracy) and MOTP (Multiple Object Tracking Precision) values for both 2D and 3D bounding boxes.
In this evaluation, a true positive is determined by a continuously tracked object across different frames whose overlapping with the ground truth is over a specific threshold.
The threshold of true positives for 2D bounding boxes is set to 0.5 and for 3D bounding boxes, the threshold is set to 0.25.

\begin{table}[h]
    \centering
    \begin{tabularx}{\linewidth}{X|XX} \hline
        & ClusterVO~\cite{clustervo} & Ours \\ \hline
        MOTP (2D) & 0.8092 & {\bf 0.8362} \\
        MOTA (2D) & 0.3359 & {\bf 0.5996} \\ \hline
        MOTP (3D) & 0.2944 & {\bf 0.7231} \\
        MOTA (3D) & -0.3563 & {\bf 0.4666} \\ \hline
    \end{tabularx}
    \caption{MOTA (Multiple Object Tracking Accuracy) and MOTP (Multiple Object Tracking Precision) evaluations on the 21 sequences of the training set in the KITTI MOT dataset.}
    \label{tab:object-tracking}
\end{table}

Since we did not find benchmark results of object tracking evaluating on the KITTI MOT dataset from existing works, we only compare our system to the state-of-the-art ClusterVO~\cite{clustervo} by running it directly.
Note that for object tracking evaluation, we use the original stereo version of~\cite{clustervo}.
From the result, we can confirm that our proposed method outperforms the original ClusterVO in both 2D and 3D.

For 2D bounding boxes, both~\cite{clustervo} and our method adopt deep object detection with high performance.
Therefore, they do not show much difference in MOTP.
However, as mentioned before,~\cite{clustervo} tracks objects based on associated low-level features, while our method adopts MOT directly using high-level 2D bounding boxes.
Hence, our method succeeds to avoid the possible instability of low-level features and outperforms~\cite{clustervo} on MOTA, which means our method can generate more true positives.
Furthermore, in the case of 3D, since our method can generate both accurate 3D bounding boxes and more true positives, our method shows a significant improvement over~\cite{clustervo}.
\cite{clustervo} tends to generate tracked objects even with a very small number of labeled features.
Hence, it creates more false positives and conversely, leads to worse performance on MOTA.
\figref{fig:object-tracking} shows a qualitative comparison between two methods.
In conclusion, we demonstrate that with monocular input and single-view depth estimation, MOTSLAM can achieve a significantly high performance compared to existing stereo methods.


\section{CONCLUSION} \label{sec:conclusion}

In this work, we present a monocular dynamic visual SLAM that simultaneously tracks poses as well as shapes of objects while performing camera tracking and mapping.
Different from the existing dynamic visual SLAM systems, which perform high-level object association after low-level feature association, we propose to perform high-level association first to provide a robust and accurate initialization.
The low-level association is then performed with the robust initialization, and both of them are optimized via bundle adjustment.

Although the current MOT algorithm~\cite{sort} is simple and effective, it only leverages 2D information and we can extend it by using 3D information including 3D bounding boxes and features to increase its performance.
Additionally, the quality of depth maps is a significant limitation of the whole system.
To tackle this problem, uncertainty maps~\cite{d3vo} can be built to filter inliers in depth maps for better performance.

\bibliographystyle{template/IEEEtran}
\bibliography{template/IEEEabrv,citations}

\end{document}